\pdfoutput=1
\documentclass[11pt]{article}

\usepackage[preprint]{acl}
\usepackage{booktabs}
\usepackage{times}
\usepackage{latexsym}
\usepackage{color}
\usepackage[T1]{fontenc}
\usepackage[utf8]{inputenc}
 \usepackage{amsmath,amssymb}
\usepackage{microtype}

\usepackage{inconsolata}
\usepackage{multicol}
\usepackage{multirow}
\usepackage{algorithm}
\usepackage[noend]{algpseudocode}

\usepackage{graphicx}
\usepackage{bbding}
\title{DReSD: Dense Retrieval for Speculative Decoding}

\author{
    Milan Gritta\textsuperscript{\normalfont1}, Huiyin Xue\thanks{\ Equal contribution with first author, conducted during a research internship at Huawei Noah's Ark Lab, London. }\textsuperscript{\normalfont2} and Gerasimos Lampouras\textsuperscript{\normalfont1} \\
    \textsuperscript{1}Huawei Noah’s Ark Lab, London, UK \\
    \textsuperscript{2}University of Sheffield, UK \\
    \texttt{\{milan.gritta,gerasimos.lampouras\}@huawei.com} \\
    \texttt{hxue12@sheffield.ac.uk}
}

\begin{document}
\maketitle
\begin{abstract}
Speculative decoding (SD) accelerates Large Language Model (LLM) generation by using an efficient draft model to propose the next few tokens, which are verified by the LLM in a single forward call, reducing latency while preserving its outputs. We focus on retrieval-based SD where the draft model retrieves the next tokens from a non-parametric datastore. Sparse retrieval \cite[REST]{he2023rest}, which operates on the surface form of strings, is currently the dominant paradigm due to its simplicity and scalability. However, its effectiveness is limited due to the usage of short contexts and exact string matching. Instead, we introduce \textbf{D}ense \textbf{Re}trieval for \textbf{S}peculative \textbf{D}ecoding (DReSD), a novel framework that uses approximate nearest neighbour search with contextualised token embeddings to retrieve the most semantically relevant token sequences for SD. Extensive experiments show that DReSD achieves (on average) 87\% higher acceptance rates, 65\% longer accepted tokens and 19\% faster generation speeds compared to sparse retrieval (REST).
\end{abstract}

\section{Introduction}
Generative transformers \cite{vaswani2017attention} are currently the dominant artificial intelligence paradigm with recent LLMs scaled to tens (or even hundreds) of billions of parameters \cite{brown2020language,liu2024deepseek,dubey2024llama}. In spite of their strong capabilities, the auto-regressive nature of generation requires a costly forward pass for each new token. Various solutions have been proposed to accelerate LLMs such as Flash Attention \cite{shah2024flashattention}, Mixture of Experts \cite{fedus2022switch,jacobs1991adaptive}, Tensor Parallelism \cite{shoeybi2019megatron}, Linear Attention \cite{qin2024lightning} and others. The focus of our work is Speculative Decoding \cite{leviathan2023fast}, which seeks to accelerate generation by using an efficient draft model to propose the next few tokens that are verified in a single forward call of the LLM \cite{stern2018blockwise}, guaranteeing its outputs.
\begin{figure}[t]
  \centering
  \includegraphics[width=1\linewidth]{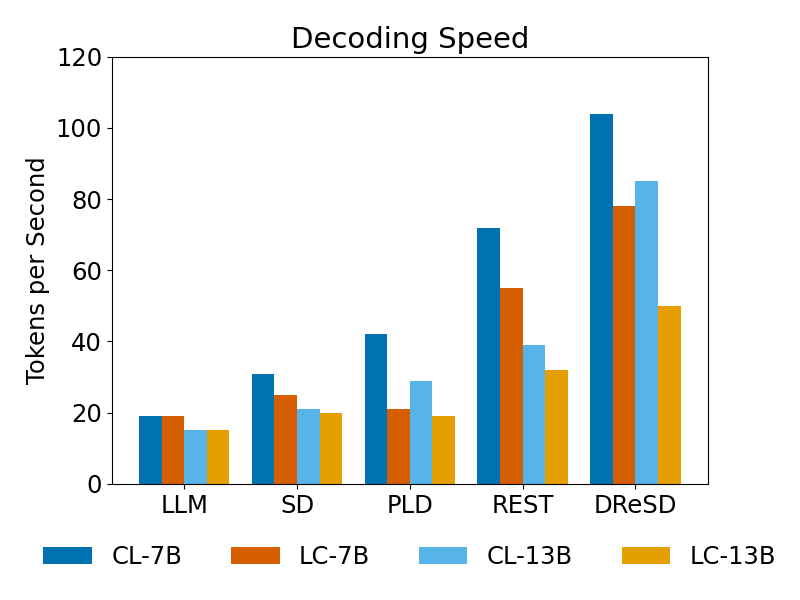}
  \caption{Fastest configurations for selected SD methods (greedy decoding), relative to auto-regressive generation (LLM), CL = CodeLlama, LC = Llama2-Chat.}
  \label{fig:tps_all}
\end{figure}
While several viable SD paradigms exist \cite{xia2024unlocking,zhang2024beyond,ryu2024closer,zimmer2024mixture}, this work specifically focuses on retrieval-based SD where a draft model retrieves token sequences from a non-parametric datastore, usually a suffix array/automaton. Sparse retrieval has established itself as the dominant paradigm \cite{he2023rest,yang2023inference,saxena2023prompt,hu2024sam}, largely due to its simplicity and efficiency. However, we hypothesise that this approach suffers from limitations such as lower precision due to the use of short contexts and reduced recall due to exact string matching. As an alternative, we introduce \textbf{D}ense \textbf{Re}trieval for \textbf{SD} (DReSD) which seeks to overcome these limitations by utilising approximate nearest neighbour search with contextualised token representations. DReSD is a novel plug-and-play SD framework based on semantic similarity that shows significantly improved acceptance rates. Through extensive experimentation, we identify three critical factors of dense retrieval for SD and show how an optimal configuration can accelerate generation by up to 4.64x\footnote{The code and data are available at \url{https://github.com/huawei-noah/HEBO/tree/DReSD}}.

\paragraph{Summary of Contributions:} We conduct a detailed comparative analysis of sparse and dense retrieval for SD in order to identify the critical factors of effective dense retrieval. To address these, we propose a novel SD framework (for the first time, to our best knowledge) for easy LLM integration. Results show that DReSD achieves (on average, across all experiments) 87\% higher acceptance rates, 65\% longer accepted tokens and 19\% faster generation compared to sparse retrieval.

\section{Background}

\subsection{Speculative Decoding}
\label{sec:sd}
Let $\mathsf{x}$ represent the input tokens ($\mathsf{x_1, x_2, ..., x_t}$) such as a prompt and any tokens generated up to time step $\mathsf{t}$. Auto-regressive generation requires a full forward pass through the model $\mathsf{x_{t+1} = LLM(x)}$ to decode every new token $\mathsf{x_{t+1}}$, which is very resource-intensive for large LLMs. Therefore, a smaller draft model $\mathcal{M}_\mathsf{DRAFT}$ efficiently proposes $\mathsf{k}$ next tokens $(\mathsf{x_{t+1}, x_{t+2}, ..., x_{t+k}})$, denoted $\mathsf{x_d}$, which can then be verified with a single forward call $\mathsf{x_v = llm\_verify(x_d)}$. Verification only accepts tokens $\mathsf{x_v}$ that would have been generated by the $\mathsf{LLM}$, irrespective of utilising SD. $\mathcal{M}_\mathsf{DRAFT}$ can be a small LLM \cite{miao2023specinfer}, a retrieval-based model \cite{he2023rest}, a subset of $\mathsf{LLM}$'s parameters called `draft heads' \cite{cai2024medusa,li2024eagle, ankner2024hydra} or no auxiliary draft model at all, called `self-drafting' \cite{mamou2024accelerating}. Each paradigm has its trade-offs and the landscape is evolving rapidly \cite{xia2024unlocking,zhang2024beyond,ryu2024closer}. 

\subsection{Retrieval-based Speculative Decoding}
Since SD operates at the token level, it requires a continuous interaction between $\mathcal{M}_\mathsf{DRAFT}$ and the $\mathsf{LLM}$. In retrieval-based SD, $\mathcal{M}_\mathsf{DRAFT}$ is represented by a non-parametric, training-free, static or dynamic datastore from which next token sequences are efficiently drafted and finally verified by the $\mathsf{LLM}$. Retrieval-based SD can be broadly divided into sparse and dense retrieval.

\subsubsection{Sparse Retrieval for SD}
Sparse retrieval employs exact string matching\footnote{\url{https://en.wikipedia.org/wiki/Suffix_array} or \url{https://en.wikipedia.org/wiki/Suffix_automaton}.} to retrieve $\mathsf{k}$ next tokens $(\mathsf{x_{t+1}, x_{t+2}, ..., x_{t+k}})$ from the datastore, which contains a large body of pre-tokenized text similar to the target task(s), allowing for appropriate drafting. There are two types of sparse retrieval datastores for SD.

\paragraph{A static datastore} keeps its content \textit{unchanged during inference}. The most similar work (and our main baseline) is Retrieval-based Speculative Decoding \cite[REST]{he2023rest}. REST matches the longest possible suffix of the current context $\mathsf{x}$, a sequence of up to $\mathsf{c}$ tokens $(\mathsf{x_{t-c}, x_{t-c+1}, ..., x_{t}})$, to exact token sequences (suffixes) in the datastore to provide $\mathsf{k}$ draft candidates $(\mathsf{x_{t+1}, x_{t+2}, ..., x_{t+k}})$ for $\mathsf{LLM}$ verification. The main limitation of exact string matching is that minor perturbations in $\mathsf{x}$ will result in a failure to retrieve useful candidate drafts.

\paragraph{A dynamic datastore} keeps updating its content \textit{continuously during inference} \cite{yang2023inference,luo2024turning,saxena2023prompt}, which means it benefits from recently generated token sequences that align well with the $\mathsf{LLM}$, particularly for tasks with repetitive texts. Combinations of static and dynamic datastores are also possible \cite{hu2024sam}. However, as the focus of our work is a systematic `apples to apples' comparison of sparse and dense retrieval, these methods are not appropriate for a direct comparison with DReSD (or REST). We aim to explore (for the first time) the comparative efficacy of static datastores for the purpose of SD.

\begin{figure*}[t]
  \centering
  \includegraphics[width=\linewidth]{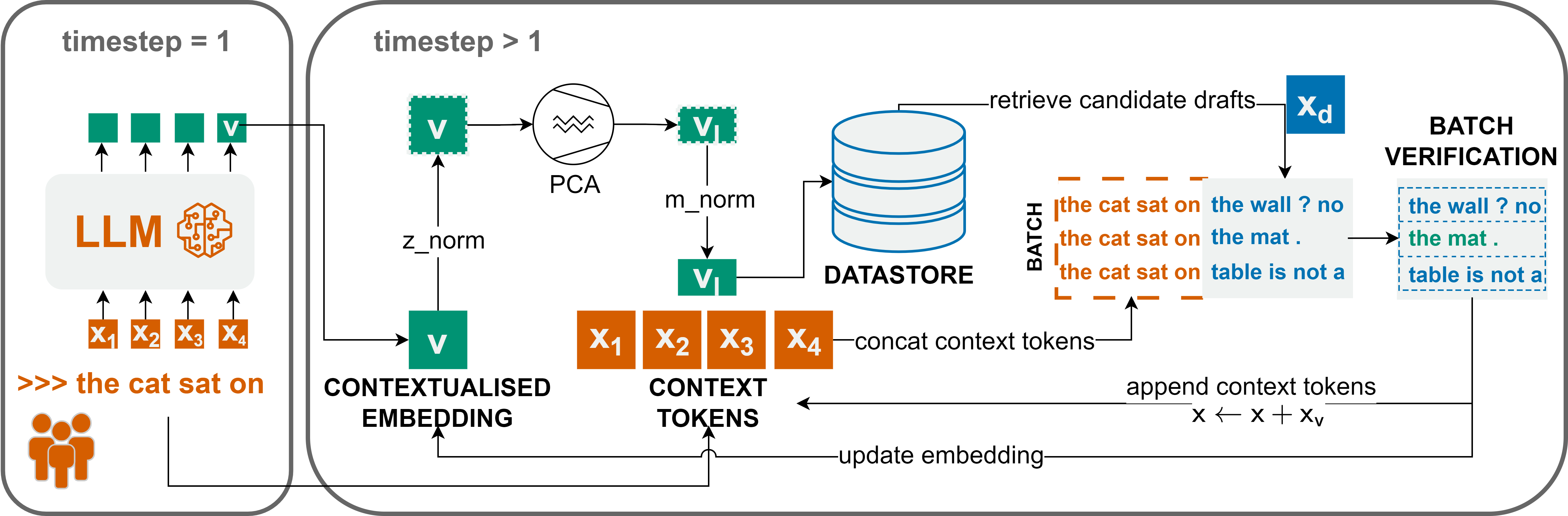}
  \caption{A flowchart of the DReSD framework.}
  \label{fig:method}
\end{figure*}

\subsection{Dense Retrieval for SD}
The key assumption behind DReSD is that semantic similarity of contextualised token embeddings should provide superior retrieval compared to exact string matching. Therefore, $\mathcal{M}_\mathsf{DRAFT}$ is represented by a non-parametric datastore that employs \textit{approximate nearest neighbour search} \cite[ANNS]{shrivastava2014asymmetric,soar_2023} to match the (full) current context $\mathsf{x}$ to similar contexts in the datastore in order to draft the next tokens $\mathsf{x_d}$ for $\mathsf{LLM}$ verification. ANNS is a technique for finding the closest data point(s) for a given query in a possibly high-dimensional vector space \cite{karpukhin2020dense}. Nearest Neighbour Speculative Decoding \cite[NEST]{li2024nearest} is the only work using dense retrieval, to our best knowledge. However, its primary focus is \textit{retrieval augmented fusion with attribution}, not SD. NEST relies on approximate verification to fuse the $\mathsf{LLM}$ and the retrieved knowledge, which means the $\mathsf{LLM}$ outputs are not guaranteed. Additionally, while NEST did not consider exact verification in their experiments, we can estimate from their results that minimal speed-ups would be achieved under that setting.

\section{DReSD}
\label{sec:dresd}

\begin{algorithm}[h]
\caption{DReSD: An algorithmic overview.}
\label{alg:sd}
\begin{algorithmic}[1]
\State $\mathsf{x} \gets \mathsf{tokenizer(prompt)}$
\State $\mathsf{v} \gets \mathsf{LLM(x)}$ \Comment{Section \ref{sec:token_emb}}.
\While{$\mathsf{not} (\mathsf{EOS} \lor \mathsf{MAX\_LEN})$}
\State $\mathsf{v} \gets \mathsf{z\_norm(v)}$ \Comment{Section \ref{sec:z-scores}.}
\State $\mathsf{v_l} \gets \mathsf{PCA(v)}$ \Comment{Section \ref{sec:pca}.}
\State $\mathsf{v_l} \gets \mathsf{m\_norm(v_l)}$ \Comment{Section \ref{sec:norm_one}.}
\State $\mathsf{x_d} \gets \mathcal{M_\mathsf{DRAFT}}\mathsf{(v_l)}$ \Comment{Section \ref{sec:db}.}
\State $\mathsf{v, x_v} \gets \mathsf{batch\_verify(x_d)}$  \Comment{Section \ref{sec:batch}.}
\State $\mathsf{x} \gets \mathsf{x + x_v}$  \Comment{Append $\mathsf{x_v}\ $}
\EndWhile
\State \Return x
\end{algorithmic}
\end{algorithm}

We are now ready to introduce \textbf{D}ense \textbf{Re}trieval for \textbf{S}peculative \textbf{D}ecoding, shown in Figure \ref{fig:method} and Algorithm \ref{alg:sd} above. Focusing on the latter, the user prompt is tokenised in step 1 and embedded in step 2. Entering the loop (3), the embedding is normalised (4), then reduced (5) to optimise storage and compute requirements. After a second normalisation step (6), we query the datastore to retrieve the draft next tokens (7). They are verified by the $\mathsf{LLM}$ (8), returning the accepted token(s) and the embedding of the last accepted token. We append the accepted token(s) to the current context and begin a new iteration, which ends when we reach maximum sequence length or the <EOS> token.

\subsection{Token Embeddings}
\label{sec:token_emb}
The initial step is to generate a contextualised token embedding $\mathsf{v} \gets \mathsf{LLM(x)}$ to represent the current state of the $\mathsf{LLM}$ that will be used to retrieve candidates for the next tokens. In DReSD, $\mathsf{v}$ is the last hidden state before the language modelling head\footnote{Alternative $\mathsf{LLM}$ components may be used for the current state representation but this is out of the scope of this work.}. As in standard SD, $\mathsf{LLM(x)}$ will also generate the next token $\mathsf{x_{t+1}}$, which we additionally use to filter retrieved candidate drafts. Even if all draft tokens are rejected, $\mathsf{x_{t+1}}$ ensures that each SD iteration produces \textit{at least one valid token}.

\subsection{Z-scores Normalisation}
\label{sec:z-scores}
Before we perform dimensionality reduction, we centre the empirical mean around 0 with a standard deviation of 1 to reduce the correlation between different embedding dimensions \cite{ethayarajh-2019-contextual,reimers-gurevych-2019-sentence}, see Equation \ref{eq:batch_norm}. We randomly sample $\sim$1 million (full size) token embeddings $\mathsf{V}$ from the datastore to estimate the mean and standard deviation for efficient inference.

\begin{equation}
  \label{eq:batch_norm}
  \mathsf{v = \frac{v - E[V]}{\sqrt{Var[V] + \epsilon}}}
\end{equation}

\subsection{Dimensionality Reduction}
\label{sec:pca}
Using the full $\mathsf{LLM}$ hidden state $\mathsf{v}$ with thousands of dimensions is not scalable. As such, data compression and noise reduction are necessary steps for DReSD to reduce storage requirements and accelerate nearest neighbour search. Principal Component Analysis \cite[PCA]{shlens2014tutorial} is a highly effective and algorithmically simple solution for this, allowing for efficient inference, too. We use PCA to transform $\mathsf{v}$ into a low-dimensional vector $\mathsf{v_l}$ that captures the largest variation in the data, using the first $\mathsf{l}$ principal components $\mathsf{W_l}$ by computing $\mathsf{v_l \gets vW_l}$. We fit the PCA model on the same $\sim$1 million token embeddings $\mathsf{V}$ from section \ref{sec:z-scores}.

\subsection{Magnitude Normalisation}
\label{sec:norm_one}
We further standardise the embedding $\mathsf{v_l}$ by scaling each to have a unit length of 1 using $L_{p}$ normalisation over the last dimension (columns), see Eq. \ref{eq:l_norm}. This is a standard transformation required for effective (dot product) nearest neighbour search.

\begin{equation}
  \label{eq:l_norm}
  \mathsf{v_l = \frac{v_l}{\max(\Vert v_l\Vert_2, \epsilon)}}
\end{equation}

\subsection{Datastore}
\label{sec:db}
We utilise Scalable Nearest Neighbours\footnote{\url{https://github.com/google-research/google-research/tree/master/scann}}\cite{avq_2020} for approximate nearest neighbour search (time complexity $\mathsf{O_{log_n}}$). The datastore $\mathcal{D}$ is formatted as a \textbf{key-value store} $f_\mathcal{D}: k \mapsto v$ where $k$ is a \textbf{token embedding} $\mathsf{v_l^t}$ at time step $\mathsf{t}$ and $v$ is a \textbf{sequence of the next $\mathsf{N}$ tokens} $\mathsf{(x_{t+1}, ..., x_{t + N})}$, obtained from datasets similar to the target task(s). Cosine similarity is used as a standard distance metric, see Equation \ref{eq:cosine}. The next token $\mathsf{x_{t+1}}$ obtained from step \ref{sec:token_emb} is used to filter drafts that do not start with $\mathsf{x_{t+1}}$, further enhancing retrieval accuracy.

\begin{equation}
    \label{eq:cosine}
    \begin{aligned}
    \mathcal{M_\mathsf{DRAFT}}(\mathsf{v_l})&=f_\mathcal{D}(\text{arg}\underset{\mathsf{v_l^t}\in\mathcal{D}}{\max}\mathsf{\ sim}(\mathsf{\mathsf{v_l}},\mathsf{v_l^t}))\\
    \mathsf{sim}(\mathsf{\mathsf{v_l}},\mathsf{v_l^t})&=\frac{\mathsf{\mathsf{v_l}}\cdot{\mathsf{\mathsf{v_l^t}}}}{\mathsf{\max(\Vert\mathsf{v_{l}}}\Vert_2\cdot\Vert\mathsf{v_l^t}\Vert_2,\epsilon)}
    \end{aligned}
\end{equation}

\subsection{Batch Verification}
\label{sec:batch}
\begin{figure}[h]
  \centering
  \includegraphics[width=0.9\linewidth]{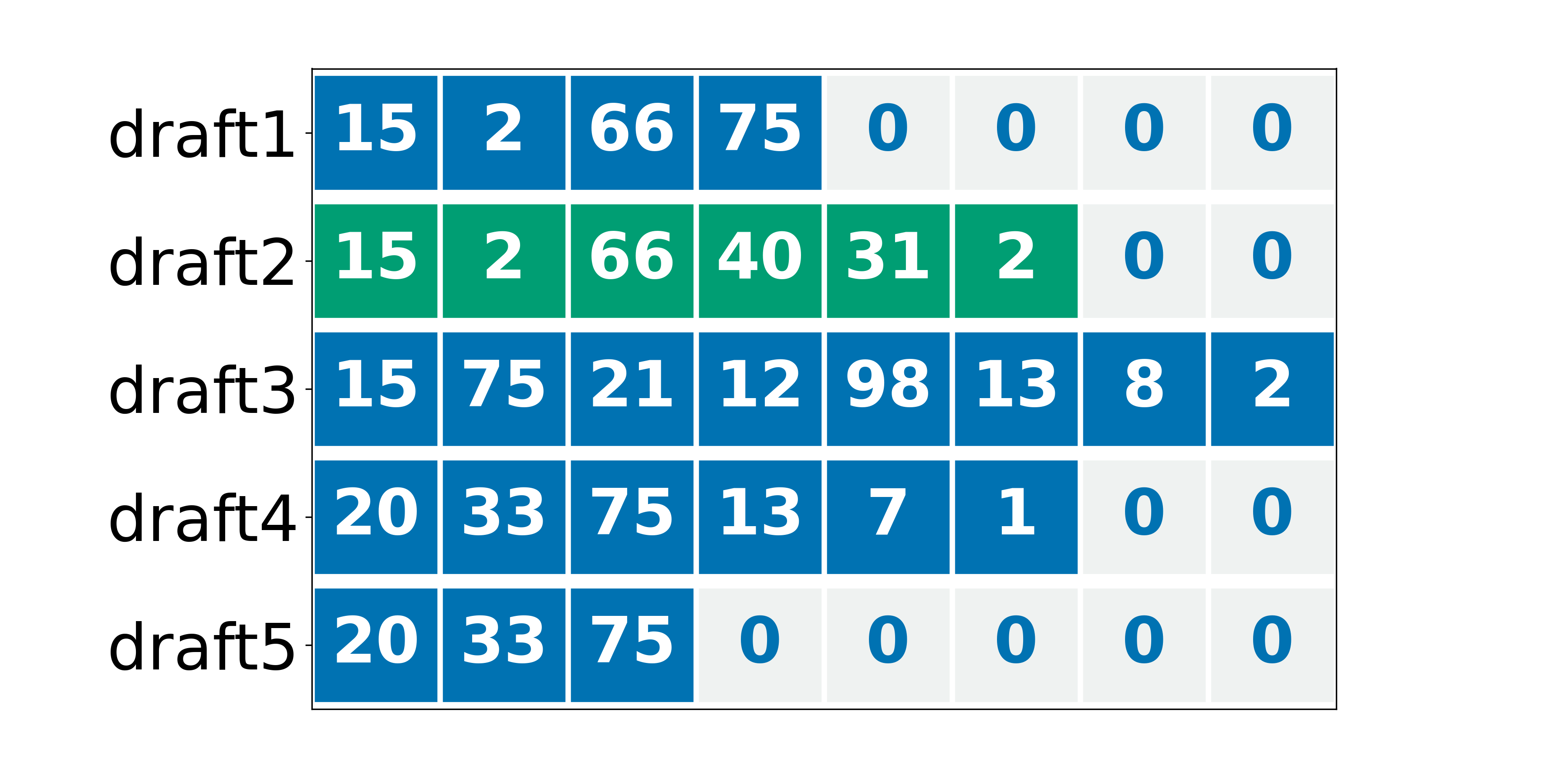}
  \caption {An illustration of batch verification with 5 drafts (rows) with a length of 8 (columns). The $\mathsf{EOS}$ id (0 in this example) is used as padding. The green sequence is accepted, blue sequences are discarded.}
  \label{fig:batch}
\end{figure}

We use batch verification \cite{yang2024multi,stewart2024n} for all experiments, which generalises standard SD verification to multiple drafts, see Figure \ref{fig:batch}. Batch verification has shown benefits for SD, particularly at lower batch sizes \cite{ni2024ems,zhang2024accelerating}. As this requires a forward call to the $\mathsf{LLM}$, we extract the embedding $\mathsf{v}$ from the last accepted token of $\mathsf{x_v}$ to efficiently feed into the next iteration (step 8, Algorithm \ref{alg:sd}). Following our baseline, for nucleus and greedy generation, we first sample tokens conditioned on the draft sequences, then accept the longest sequence that \textbf{exactly matches the outputs} of the $\mathsf{LLM}$. 

\section{Experimental Setup}

\subsection{Models}
We evaluate methods on LLMs from the Llama2
family \cite{touvron2023llama}, courtesy of Huggingface transformers \cite{wolf2019huggingface}. Specifically, we benchmark CodeLlama (7B and 13B), CodeLlama-Instruct (7B) and Llama2-Chat (7B and 13B). The $\mathcal{M_\mathsf{DRAFT}}$ for vanilla Speculative Decoding (with a small LLM drafter) features Llama-Chat-68M\footnote{\url{https://huggingface.co/Felladrin/Llama-68M-Chat-v1}}, fine-tuned from Llama-68M \cite{miao2023specinfer}.

\subsection{Datasets and Tasks}
\label{sec:datasets}
We test models on 100 randomly selected CodeAlpaca \cite{codealpaca} prompts, which include code generation, debugging, explanation and other code tasks. The datastore for this code assistant is built from EvolInstructCode \cite{luo2023wizardcoder}, comprising $\sim$78K prompts with responses, truncated to 1,024 max tokens. We also evaluate on 80 MT-Bench\footnote{\url{https://huggingface.co/datasets/HuggingFaceH4/mt_bench_prompts}} \cite{zheng2023judging} prompts (first turn specifically, due to the compute required for the number of experiments). The datastore for this general personal assistant is built from a random subset of 80K (`train-sft') UltraChat-200K\footnote{\url{https://huggingface.co/datasets/HuggingFaceH4/ultrachat_200k}} \cite{ding2023enhancing} examples, prompts and responses truncated to 1,024 max tokens, once again, first turn to limit the scope of the long, multi-turn conversations. For the final datastore sizes, see Table \ref{table:sizes}.

\begin{table}[h]
\centering
 \begin{tabular}{lcccc}
 \toprule
 Models & EVOL & MRR & U-CHAT & MRR \\
 \midrule
  \multicolumn{5}{c}{OOD datastores (Sec. \ref{sec:in_distro})} \\ \midrule
 CL-7B & 30.9M & 93.7 & 46.3M & 97.5 \\
 CL-13B & 30.9M & 92.5 & - & - \\
 LC-7B & 30.9M & 93.6 & 46.3M & 97.9 \\
 LC-13B & 30.9M & 93.7 & - & - \\ 
 CL-I-7B & 30.9 & 93.9 & 46.3M & 97.9 \\ \midrule
 \multicolumn{5}{c}{Sec. ID datastores (Sec. \ref{sec:in_distro})} \\ \midrule
 CL-7B & 19.3M & 87.6 & - & - \\
 CL-13B & 19.3M & 85.5 & - & - \\
 LC-7B & 19M & 90.3 & 56.8M & 75.2 \\
 LC-13B & 19M & 90.4 & - & - \\
 CL-I-7B & - & - & 57.2M & 75.9 \\ \bottomrule
 \end{tabular}
 \caption{Datastore sizes in tokens + corresponding MRR. EVOL = EvolInstructCode, U-CHAT = UltraChat.}
 \label{table:sizes}
\end{table}

\subsubsection{In-Distribution Data}
\label{sec:in_distro}
The datasets used to populate the datastore are often generated by some version of ChatGPT whose outputs are not necessarily representative of the target $\mathsf{LLM}$ (Llama2). That is, the datastore outputs are out-of-distribution (OOD) with respect to the $\mathsf{LLM}$. As this divergence increases, the acceptance rates and decoding speeds are expected to decrease. To investigate the effect of in-distribution (ID) token sequences, we generate responses for each $\mathsf{LLM}$ and use those to populate the datastore. We refer to Llama2 responses as the \textbf{ID datastore} and ChatGPT responses as the \textbf{OOD datastore}.

\subsection{Metrics}
\label{sec:metrics}

\paragraph{Hardware Dependent} metrics are heavily influenced by the choice, availability and optimisation level of hardware components. Nevertheless, in order to provide indicative walltime improvements, we use \textbf{tokens-per-second} (abbreviated to TPS) as the standard metric, reporting the median of three runs. TPS is measured on a single NVIDIA V100 (32GB) GPU with 96 CPUs and 500GB of RAM.

\paragraph{Hardware Independent} metrics are more appropriate for algorithmic comparisons that are independent of optimisation tricks and hardware quality. \textbf{Mean Acceptance Rate} (\textbf{MAR}) is the number of tokens drafted divided by the number of tokens accepted by the $\mathsf{LLM}$. MAR is computed at the prompt level, then averaged over all prompts.

\paragraph{Retrieval Only} We also conduct intrinsic evaluation to assess the quality of the nearest neighbour search. We use Mean Reciprocal Rank\footnote{A presence of duplicate embeddings in a large datastore can lead to lower scores, even with near-perfect retrieval.}, shown in Equation \ref{eq:mrr}, where $\mathsf{rank_i}$ is the position of the correct item and $\mathsf{N}$ is the number of embeddings in the datastore (a score of 1 equates to perfect retrieval).

\begin{equation}
  \label{eq:mrr}
  \mathsf{MRR = \frac{1}{N}\sum_{i=1}^{N}\frac{1}{rank_i}}
\end{equation}

\section{Results}

We provide reference metrics for auto-regressive decoding with the base $\mathsf{LLM}$, vanilla speculative decoding \cite{leviathan2023fast} and Prompt Lookup Decoding (PLD), a dynamic retrieval method that uses the current input tokens for drafting \cite{saxena2023prompt}. Our main point of reference is REST \cite{he2023rest}, which is created from the same data as DReSD. For all experiments, we generate up to 128 new tokens per prompt.

\subsection{Mean Acceptance Rates}
\label{sec:acceptance}
\begin{figure}[h]
  \centering
  \includegraphics[width=0.49\textwidth]{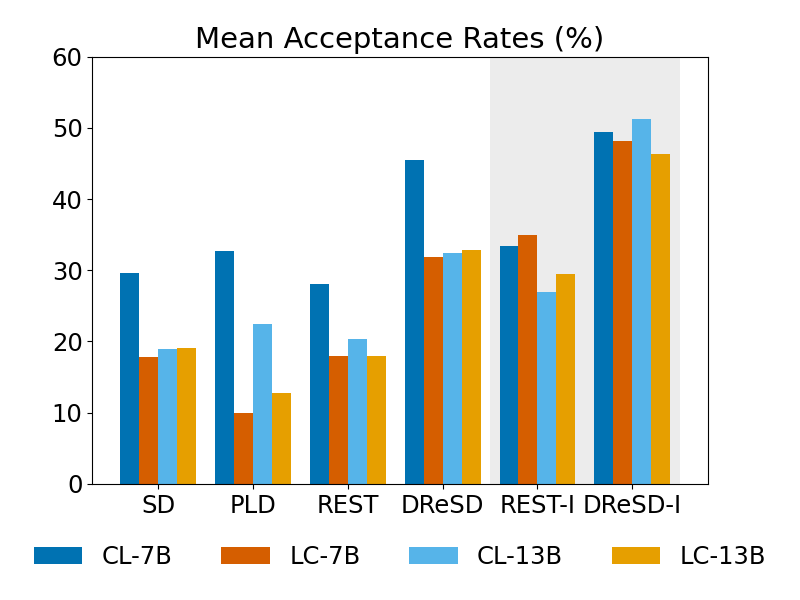}
  \caption{Mean Acceptance Rates (MAR) for the Code Assistant. Suffix  "-I" denotes the ID datastore setting.}
  \label{fig:mar}
\end{figure}

We first examine the average acceptance rates in a highly controlled setting where the draft lengths are as identical as possible. This is to test our core assumption that, all things being equal, dense retrieval would more accurately match the current context to useful sequences of next tokens in the datastore, relative to sparse retrieval. The hypothesis is confirmed in Figure \ref{fig:mar} as significantly higher MAR for DReSD, on average 87\% higher. This translates to fewer verification calls due to longer accepted drafts, on average 65\% longer than REST. 

\subsection{Effective Dense Retrieval}
\label{sec:retrieval}
\begin{figure}[t]
  \centering
  \includegraphics[width=0.95\linewidth]{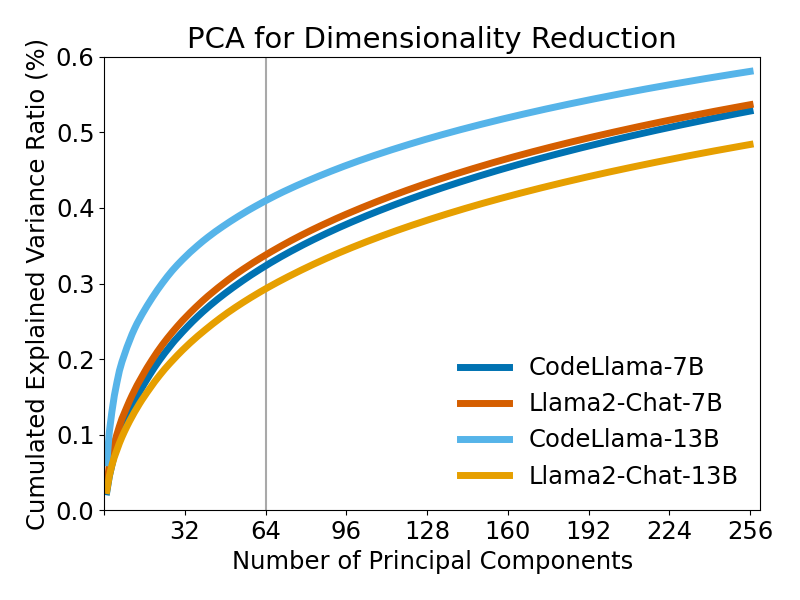}
  \caption{Cumulative Explained Variance Ratio for a 256-dimensional PCA model. We use the first 64 dims.}
  \label{fig:pca}
\end{figure}

Sparse retrieval libraries had been highly optimised over time, therefore, a high-performing datastore is a critical component of DReSD. It is imperative to maximise the algorithmic efficiency of DReSD to amortise the relatively higher cost of ``vanilla'' dense retrieval. Reducing the large dimensionality\footnote{\url{https://pypi.org/project/torch-pca/}} of $\mathsf{LLM}$ hidden states while preserving the most informative features is a top priority. Figure \ref{fig:pca} shows the cumulative explained variance ratio for the first 256 principal components from which we select 64 after a dimensionality ablation. Table \ref{table:pca_ablate} shows that retaining more than 64 principal components provides no meaningful improvement in downstream metrics. The pre-PCA (Equation \ref{eq:batch_norm}) and post-PCA (Equation \ref{eq:l_norm}) normalisation steps are particularly important since the MRR scores sharply drop without these transformations. The most remarkable observation is that selecting just over 1\% of the 4096 to 5120 dimensional $\mathsf{LLM}$ hidden state features is enough to capture between 30\%-40\% of explained variance in PCA, and achieves such strong retrieval performance (see MRR, in Table \ref{table:sizes} and \ref{table:pca_ablate}). There is a surprisingly high degree of redundancy in the $\mathsf{LLM}$ hidden state in terms of the minimum features required for effective dense retrieval, an important discovery for any future work.

\begin{table}[h]
\centering
 \begin{tabular}{lcccc}
 \toprule
 \textbf{Metrics} & \textbf{32D} & \textbf{64D} & \textbf{96D} & \textbf{128D} \\ \midrule
 MAR & 20.1 & 22.5 & 22.8 & 23 \\ 
 Calls & 37 & 34.6 & 34.7 & 34.5 \\ 
 TPS & 32 & 35 & 36 & 35 \\ \midrule
 \textbf{MRR} & \textbf{85.5} & \textbf{93.6} & \textbf{94.2} & \textbf{94.2} \\ \bottomrule
 \end{tabular}
 \caption{An ablation of PCA dimensionality reduction. `Calls' = the average number of $\mathsf{LLM}$ verification calls.}
 \label{table:pca_ablate}
\end{table}

\begin{figure}[t]
  \centering
  \includegraphics[width=0.95\linewidth]{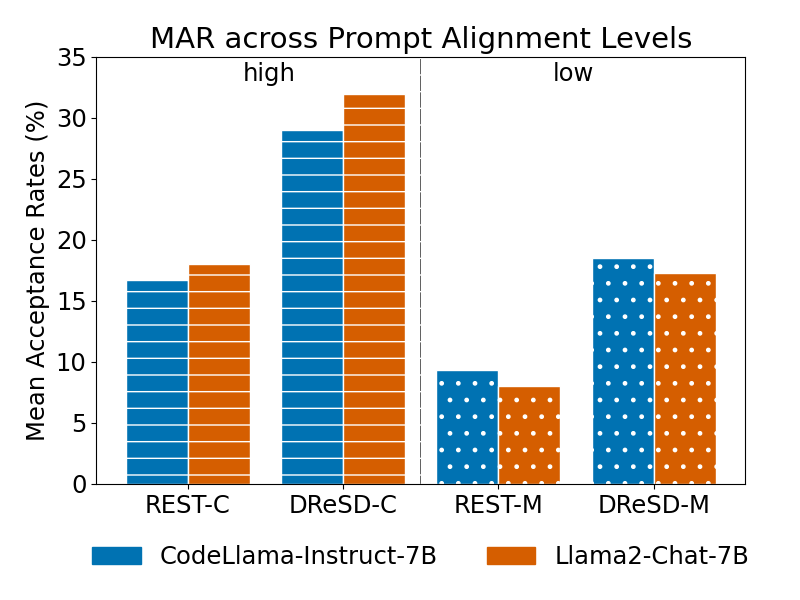}
  \caption{Mean Acceptance Rates with high (CodeAlpaca, "-C") \& low (MT-Bench, "-M") prompt alignment.}
  \label{fig:mar_instruct}
\end{figure}

\subsection{Importance of Datastore Alignment}
\label{sec:in_distro_results}
Another critical component of retrieval-based SD is \textit{datastore alignment}, which we split into a) \textbf{prompt alignment}, b) \textbf{response alignment}, and c) \textbf{sampling alignment}. These \textit{multiplicatively} influence overall effectiveness, which means that poor alignment in any of them can adversely impact performance. Let us examine why in more detail.

\begin{figure*}[t]
  \centering
  \includegraphics[width=1\linewidth]{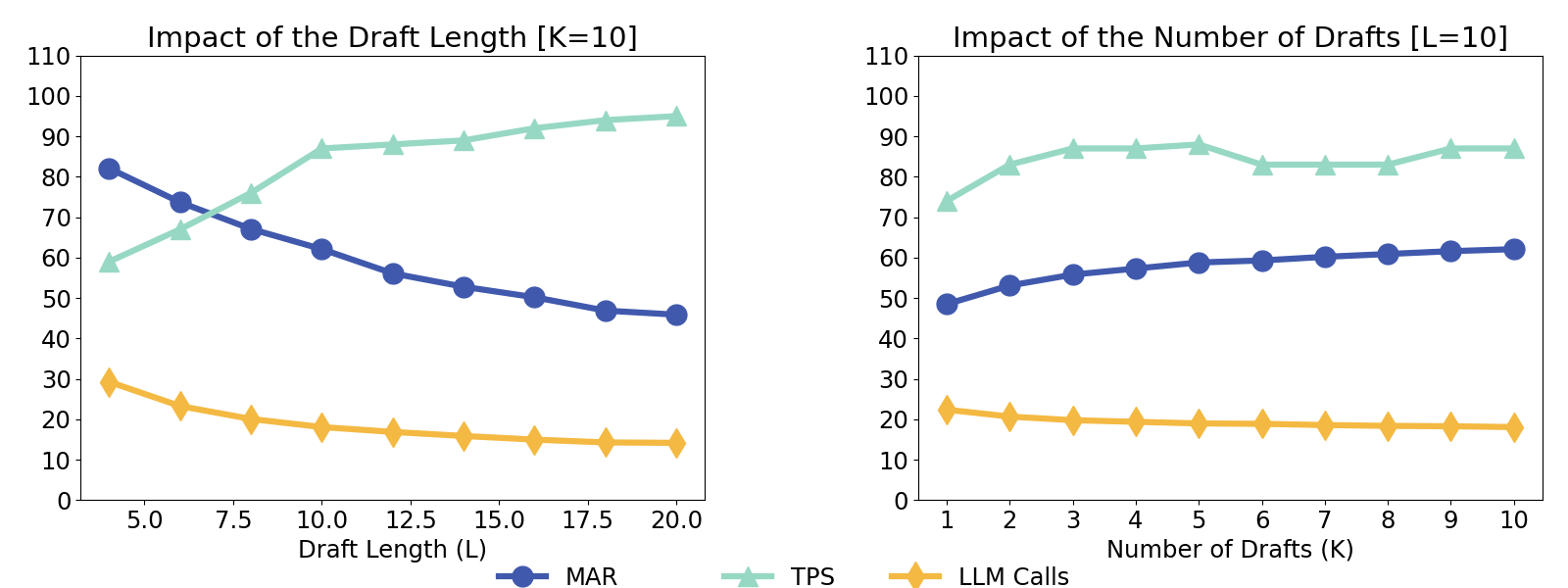}
  \caption{Investigating the impact of increasing draft lengths (left) and the number of drafts (right) on MAR, TPS and LLM verification calls, using a greedy datastore with greedy generation on code assistant tasks (CodeAlpaca).}
  \label{fig:len}
\end{figure*}

\begin{figure}[t]
  \centering
  \includegraphics[width=1\linewidth]{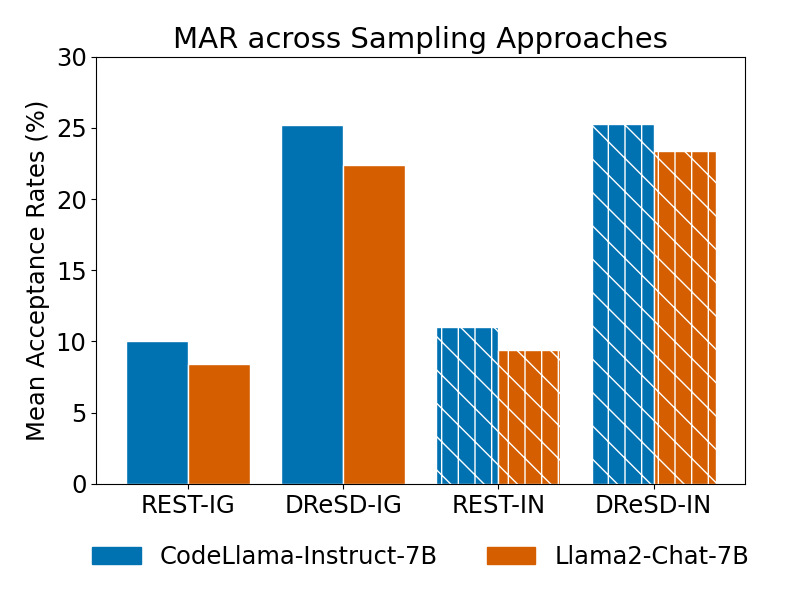}
  \caption{MAR for "-I" = ID datastore (nucleus generation) with "-N" = nucleus and "-G" greedy sampling.}
  \label{fig:mat_instruct}
\end{figure}

\paragraph{Prompt alignment} is typically satisfied by populating the datastore with prompts highly related to the target task(s) such as code, maths or general question answering. After a brief qualitative assessment, this appears to have been reasonably satisfied for the code assistant, however, only to some degree for MT-Bench tasks. Despite strong response alignment (ID datastore) and retrieval (MRR, Table \ref{table:sizes}), the poor prompt alignment leads to lower acceptance rates (see Figure \ref{fig:mar_instruct}), relative to CodeAlpaca. Since this result reflects prior findings \cite{he2023rest}, we think that choosing a more prompt-aligned dataset will result in faster decoding. In any case, dense retrieval (DReSD) outperforms sparse retrieval (REST) by $\sim$90\% in Figure \ref{fig:mar_instruct}.

\begin{figure*}[t]
  \centering
  \includegraphics[width=1\linewidth]{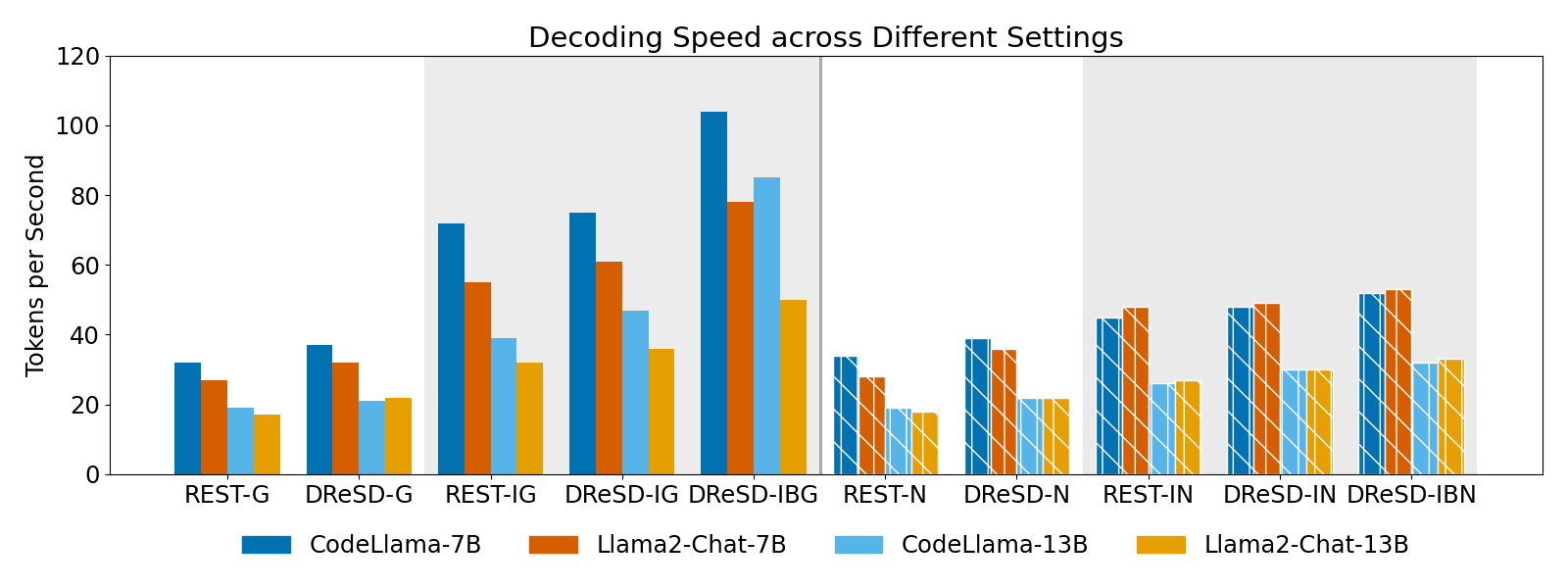}
  \caption{TPS metrics for a selection of LLMs and configs: "-G" = greedy decoding, "-I" = uses ID datastore (greedy generation), "-N" = nucleus sampling, "-B" = our best setup (see section \ref{sec:walltime}), LLM = auto-regressive generation.}
  \label{fig:tps_chart}
\end{figure*}

\paragraph{Response alignment} is the similarity of outputs between the $\mathsf{LLM}$ and the model(s) that generated the datastore, i.e. draft sequences with a low probability under the $\mathsf{LLM}$ lead to high rejection rates and slow decoding speeds, regardless of the capabilities of the model(s) that generated the datastore. A qualitative comparison of Llama2 (ID datastore) and ChatGPT (OOD datastore) responses reveals significant differences in writing styles, knowledge depth and response lengths. The effects of these differences can be observed in Figures \ref{fig:tps_all}, \ref{fig:mar} and \ref{fig:tps_chart} by comparing methods with and without the suffix "-I", showing that the ID datastore has a strong positive effect in all cases. For example, MAR increased by $\sim$70\% on average for CodeAlpaca with an ID datastore despite being $\sim$40\% smaller than the OOD datastore, emphasising the importance of response alignment over sheer data quantity. 

\paragraph{Sampling alignment} refers to the similarity of hyperparameters with which the datastore content was generated and the sampling hyperparameters at inference time. For instance, in Figure \ref{fig:tps_chart} (left), the best speed-ups were achieved with greedy sampling and a `greedy' datastore. In contrast, nucleus sampling (temperature=0.7, p=0.95\footnote{Nucleus hyperparameters are the same in all experiments.}) with a `greedy' datastore resulted in lower speed-ups (Figure \ref{fig:tps_chart}, right). In another ablation (Figure \ref{fig:mat_instruct}), the ID datastore was generated with nucleus sampling, using 3 responses of up to 128 tokens per prompt (see Table \ref{table:sizes} for final datastore sizes). There was no significant difference between MAR scores with greedy or nucleus sampling this time. In summary, the more permissive the sampling parameters are, e.g. a high temperature, the greater the LLM's expressivity that will need to be covered by the datastore. This is particularly the case for open-ended tasks such as creative writing where the number of `correct' responses is usually much greater than in a coding or maths task, for example. In contrast, low temperature sampling can achieve fast decoding speeds with a much smaller `greedy' datastore.

\subsection{Optimal Draft Shape}
\label{sec:draft_lengths}
The final critical factor for achieving high decoding performance is the shape of the draft because each additional token sent for verification increases the cost of the $\mathsf{LLM}$ forward pass (see Figure \ref{fig:batch} for an illustration of a 5 x 8 draft). While REST outputs drafts that encode shallow and wide trees (10 x 6, on average, cannot be altered), DReSD can modify this shape to potentially achieve higher speed-ups. For instance, when the draft shape is matched to REST, DReSD outperforms it by 15\% to 28\% with nucleus sampling and 10\% to 29\% with greedy decoding (TPS, Figure \ref{fig:tps_chart}). Switching to the ID datastore (suffix "-I"), the margins are slightly smaller, 2\% to 15\% for nucleus sampling and 4\% to 20\% for greedy generation. However, the optimal REST draft (wide and shallow) is not necessarily optimal for DReSD (narrow and deep). Therefore, we investigate the optimal draft shape using an ablation of \textbf{the number and the length of drafts} with a 'greedy' ID datastore and greedy decoding, shown in Figure \ref{fig:len}. The number of drafts is fixed to 10 for `Impact of Draft Length' (left) and the length of each draft is fixed to 10 tokens for `Impact of the Number of Drafts' (right). Based on the findings of this ablation, the best configurations are DReSD-IBN (10 x 10), yielding between 10\% and 23\% speed-up relative to REST for nucleus sampling and DReSD-IBG, (3 x 20), yielding 42\% to 218\% faster speeds for greedy decoding. As a general principle: 1) we should increase the draft length for higher acceptance rates and vice versa, and 2) include fewer drafts for greedy generation but more drafts (shorter) for nucleus sampling.

\begin{table}[h]
\centering
 \begin{tabular}{lcccc}
 \toprule
 Models & SD & PLD & REST & DReSD \\ \midrule
 CL-7B & 1.63x & 2.21x & 3.79x & \textbf{5.47x} \\ 
 LC-7B & 1.32x & 1.11x & 2.89x & \textbf{4.11x} \\
 CL-13B & 1.40x & 1.93x & 2.60x & \textbf{5.67x} \\
 LC-13B & 1.33x & 1.27x & 2.13x & \textbf{3.33x} \\ \midrule
 Average & 1.42x & 1.63x & 2.85x & \textbf{4.64x} \\ \bottomrule
 \end{tabular}
 \caption{Average speed-ups relative to auto-regressive generation on the code assistant tasks (CodeAlpaca).}
 \label{table:speedup}
\end{table}

\subsection{Walltime Improvements}
\label{sec:walltime}

Table \ref{table:speedup} provides the speed-up ratios for vanilla SD with Llama-Chat-68M, Prompt Lookup Decoding (dynamic sparse retrieval), REST (static sparse retrieval) and DReSD (static dense retrieval). The averages show that every SD method accelerates standard auto-regressive $\mathsf{LLM}$ generation by at least 40\%, with higher speed-ups observed for the 'smaller' 7B models. PLD is most effective for very repetitive outputs, which can be a property of the task and/or the model. For example, CodeLlama (not instruction-tuned), has a tendency to produce repetitive texts, particularly on code assistant tasks. This is why the effectiveness of PLD drops sharply for instruction-tuned models. Our best configuration for REST, shown in Figure \ref{fig:tps_chart} with suffix "-IG", gives an average 2.85x speed-up. DReSD using drafts of shape (3 x 20 tokens) and the ID datastore (suffix "-IBG") averaged a remarkable 4.64x improvement over auto-regressive decoding. Figure \ref{fig:tps_all} provides a visual summary of these configurations in relation to other baseline methods. The speed-ups on MT-Bench are relatively more modest, up to $\sim$1.52x, due to poor prompt alignment discussed earlier (\ref{sec:in_distro_results}). Still, DReSD significantly outperformed REST, which achieved only $\sim$1.15x speed-up with the same dataset(s). This confirms our hypothesis that dense retrieval is the superior search paradigm for speculative decoding.

\subsection{Storage Requirements}
Compared to sparse retrieval, the datastore size for dense retrieval is significantly larger (even with 64-dimensional embeddings), on average between 30x-40x, depending on the dataset/task and tokenizer. For example, the datastore size for MT-Bench using Llama-2-Chat-7B is only 15.7GB (46.3 million tokens). CodeAlpaca (19 million tokens) is even smaller at just 6.49GB for an average of 4.64x acceleration over baseline (Table \ref{table:speedup}). Given that disk storage and RAM are significantly cheaper than GPU rental, the running costs of DReSD are expected to be lower than REST, despite the datastore overheads. Lastly, the computationally demanding datastore creation is a \textit{one-time operation}.

\section{Conclusions}
We have presented a comparative analysis of dense and sparse retrieval for speculative decoding in order to identify and overcome the limitations of the dominant (sparse) paradigm. To address these, we have introduced DReSD, \textbf{D}ense \textbf{Re}trieval for \textbf{S}peculative \textbf{D}ecoding, a novel framework that retrieves candidate drafts from a non-parametric datastore based on semantic similarity (via approximate nearest neighbour search) instead of exact string matching. DReSD introduces a scalable and effective dense retrieval protocol that can easily integrate into modern LLMs. Exhaustive comparisons using several model and task configurations have demonstrated that DReSD achieves (on average across all settings) 87\% higher acceptance rates, 65\% longer accepted tokens and 19\% faster generation speeds compared to sparse retrieval (REST). This is enabled by three critical factors: a) a fast and accurate dense retrieval via dimensionality reduction and dual normalisation of $\mathsf{LLM}$ embeddings, b) a careful datastore alignment (particularly the ID datastore) with high acceptance rates, longer drafts and fewer $\mathsf{LLM}$ calls, c) an optimal draft shape explored via careful ablations that enabled up to 4.64x average speed-ups over baseline auto-regressive generation. We hope that our findings will enable new retrieval-based SD methodologies in the future.

\section{Limitations}

We acknowledge that sparse retrieval methods typically have lower storage and preprocessing requirements, which can make their adoption more feasible for low compute budgets compared to our proposed methodology. Simultaneously, we recognise the lack of software and/or hardware optimisation for DReSD that could fully realise its potential in terms of faster decoding speeds compared to the highly optimised sparse retrieval libraries. Finally, related work has shown that combinations of dynamic and static retrieval may bring complementary strengths to the overall approach, therefore, DReSD could be extended to such hybrid speculative decoding version in future work. In terms of social impact or ethical concerns, our work has no particular impact as we solely focus our methodology on decoding speeds from pretrained language models. No modification of datasets and/or LLMs has been performed in this work. For full transparency, all code, models and data are readily available in the open-source NLP community.

% \section*{Acknowledgments}

\bibliography{custom}

\begin{thebibliography}{43}
\providecommand{\natexlab}[1]{#1}

\bibitem[{Ankner et~al.(2024)Ankner, Parthasarathy, Nrusimha, Rinard, Ragan-Kelley, and Brandon}]{ankner2024hydra}
Zachary Ankner, Rishab Parthasarathy, Aniruddha Nrusimha, Christopher Rinard, Jonathan Ragan-Kelley, and William Brandon. 2024.
\newblock Hydra: Sequentially-dependent draft heads for medusa decoding.
\newblock \emph{arXiv preprint arXiv:2402.05109}.

\bibitem[{Brown et~al.(2020)Brown, Mann, Ryder, Subbiah, Kaplan, Dhariwal, Neelakantan, Shyam, Sastry, Askell et~al.}]{brown2020language}
Tom Brown, Benjamin Mann, Nick Ryder, Melanie Subbiah, Jared~D Kaplan, Prafulla Dhariwal, Arvind Neelakantan, Pranav Shyam, Girish Sastry, Amanda Askell, et~al. 2020.
\newblock Language models are few-shot learners.
\newblock \emph{Advances in neural information processing systems}, 33:1877--1901.

\bibitem[{Cai et~al.(2024)Cai, Li, Geng, Peng, Lee, Chen, and Dao}]{cai2024medusa}
Tianle Cai, Yuhong Li, Zhengyang Geng, Hongwu Peng, Jason~D Lee, Deming Chen, and Tri Dao. 2024.
\newblock Medusa: Simple llm inference acceleration framework with multiple decoding heads.
\newblock \emph{arXiv preprint arXiv:2401.10774}.

\bibitem[{Chaudhary(2023)}]{codealpaca}
Sahil Chaudhary. 2023.
\newblock Code alpaca: An instruction-following llama model for code generation.
\newblock \url{https://github.com/sahil280114/codealpaca}.

\bibitem[{Ding et~al.(2023)Ding, Chen, Xu, Qin, Zheng, Hu, Liu, Sun, and Zhou}]{ding2023enhancing}
Ning Ding, Yulin Chen, Bokai Xu, Yujia Qin, Zhi Zheng, Shengding Hu, Zhiyuan Liu, Maosong Sun, and Bowen Zhou. 2023.
\newblock \href {https://arxiv.org/abs/2305.14233} {Enhancing chat language models by scaling high-quality instructional conversations}.
\newblock \emph{Preprint}, arXiv:2305.14233.

\bibitem[{Dubey et~al.(2024)Dubey, Jauhri, Pandey, Kadian, Al-Dahle, Letman, Mathur, Schelten, Yang, Fan et~al.}]{dubey2024llama}
Abhimanyu Dubey, Abhinav Jauhri, Abhinav Pandey, Abhishek Kadian, Ahmad Al-Dahle, Aiesha Letman, Akhil Mathur, Alan Schelten, Amy Yang, Angela Fan, et~al. 2024.
\newblock The llama 3 herd of models.
\newblock \emph{arXiv preprint arXiv:2407.21783}.

\bibitem[{Ethayarajh(2019)}]{ethayarajh-2019-contextual}
Kawin Ethayarajh. 2019.
\newblock \href {https://doi.org/10.18653/v1/D19-1006} {How contextual are contextualized word representations? {C}omparing the geometry of {BERT}, {ELM}o, and {GPT}-2 embeddings}.
\newblock In \emph{Proceedings of the 2019 Conference on Empirical Methods in Natural Language Processing and the 9th International Joint Conference on Natural Language Processing (EMNLP-IJCNLP)}, pages 55--65, Hong Kong, China. Association for Computational Linguistics.

\bibitem[{Fedus et~al.(2022)Fedus, Zoph, and Shazeer}]{fedus2022switch}
William Fedus, Barret Zoph, and Noam Shazeer. 2022.
\newblock Switch transformers: Scaling to trillion parameter models with simple and efficient sparsity.
\newblock \emph{Journal of Machine Learning Research}, 23(120):1--39.

\bibitem[{Guo et~al.(2020)Guo, Sun, Lindgren, Geng, Simcha, Chern, and Kumar}]{avq_2020}
Ruiqi Guo, Philip Sun, Erik Lindgren, Quan Geng, David Simcha, Felix Chern, and Sanjiv Kumar. 2020.
\newblock \href {https://arxiv.org/abs/1908.10396} {Accelerating large-scale inference with anisotropic vector quantization}.
\newblock In \emph{International Conference on Machine Learning}.

\bibitem[{He et~al.(2023)He, Zhong, Cai, Lee, and He}]{he2023rest}
Zhenyu He, Zexuan Zhong, Tianle Cai, Jason~D Lee, and Di~He. 2023.
\newblock Rest: Retrieval-based speculative decoding.
\newblock \emph{arXiv preprint arXiv:2311.08252}.

\bibitem[{Hu et~al.(2024)Hu, Wang, Zhang, Li, and Chen}]{hu2024sam}
Yuxuan Hu, Ke~Wang, Jing Zhang, Cuiping Li, and Hong Chen. 2024.
\newblock Sam decoding: Speculative decoding via suffix automaton.
\newblock \emph{arXiv preprint arXiv:2411.10666}.

\bibitem[{Jacobs et~al.(1991)Jacobs, Jordan, Nowlan, and Hinton}]{jacobs1991adaptive}
Robert~A Jacobs, Michael~I Jordan, Steven~J Nowlan, and Geoffrey~E Hinton. 1991.
\newblock Adaptive mixtures of local experts.
\newblock \emph{Neural computation}, 3(1):79--87.

\bibitem[{Karpukhin et~al.(2020)Karpukhin, O{\u{g}}uz, Min, Lewis, Wu, Edunov, Chen, and Yih}]{karpukhin2020dense}
Vladimir Karpukhin, Barlas O{\u{g}}uz, Sewon Min, Patrick Lewis, Ledell Wu, Sergey Edunov, Danqi Chen, and Wen-tau Yih. 2020.
\newblock Dense passage retrieval for open-domain question answering.
\newblock \emph{arXiv preprint arXiv:2004.04906}.

\bibitem[{Leviathan et~al.(2023)Leviathan, Kalman, and Matias}]{leviathan2023fast}
Yaniv Leviathan, Matan Kalman, and Yossi Matias. 2023.
\newblock Fast inference from transformers via speculative decoding.
\newblock In \emph{International Conference on Machine Learning}, pages 19274--19286. PMLR.

\bibitem[{Li et~al.(2024{\natexlab{a}})Li, Chen, Holtzman, Chen, Lin, Yih, and Lin}]{li2024nearest}
Minghan Li, Xilun Chen, Ari Holtzman, Beidi Chen, Jimmy Lin, Wen-tau Yih, and Xi~Victoria Lin. 2024{\natexlab{a}}.
\newblock Nearest neighbor speculative decoding for llm generation and attribution.
\newblock \emph{arXiv preprint arXiv:2405.19325}.

\bibitem[{Li et~al.(2024{\natexlab{b}})Li, Wei, Zhang, and Zhang}]{li2024eagle}
Yuhui Li, Fangyun Wei, Chao Zhang, and Hongyang Zhang. 2024{\natexlab{b}}.
\newblock Eagle: Speculative sampling requires rethinking feature uncertainty.
\newblock \emph{arXiv preprint arXiv:2401.15077}.

\bibitem[{Liu et~al.(2024)Liu, Feng, Xue, Wang, Wu, Lu, Zhao, Deng, Zhang, Ruan et~al.}]{liu2024deepseek}
Aixin Liu, Bei Feng, Bing Xue, Bingxuan Wang, Bochao Wu, Chengda Lu, Chenggang Zhao, Chengqi Deng, Chenyu Zhang, Chong Ruan, et~al. 2024.
\newblock Deepseek-v3 technical report.
\newblock \emph{arXiv preprint arXiv:2412.19437}.

\bibitem[{Luo et~al.(2024)Luo, Wang, Zhu, Zhang, Zhang, Yang, Xu, and Che}]{luo2024turning}
Xianzhen Luo, Yixuan Wang, Qingfu Zhu, Zhiming Zhang, Xuanyu Zhang, Qing Yang, Dongliang Xu, and Wanxiang Che. 2024.
\newblock Turning trash into treasure: Accelerating inference of large language models with token recycling.
\newblock \emph{arXiv preprint arXiv:2408.08696}.

\bibitem[{Luo et~al.(2023)Luo, Xu, Zhao, Sun, Geng, Hu, Tao, Ma, Lin, and Jiang}]{luo2023wizardcoder}
Ziyang Luo, Can Xu, Pu~Zhao, Qingfeng Sun, Xiubo Geng, Wenxiang Hu, Chongyang Tao, Jing Ma, Qingwei Lin, and Daxin Jiang. 2023.
\newblock Wizardcoder: Empowering code large language models with evol-instruct.
\newblock \emph{arXiv preprint arXiv:2306.08568}.

\bibitem[{Mamou et~al.(2024)Mamou, Pereg, Korat, Berchansky, Timor, Wasserblat, and Schwartz}]{mamou2024accelerating}
Jonathan Mamou, Oren Pereg, Daniel Korat, Moshe Berchansky, Nadav Timor, Moshe Wasserblat, and Roy Schwartz. 2024.
\newblock Accelerating speculative decoding using dynamic speculation length.
\newblock \emph{arXiv preprint arXiv:2405.04304}.

\bibitem[{Miao et~al.(2023)Miao, Oliaro, Zhang, Cheng, Wang, Zhang, Wong, Zhu, Yang, Shi et~al.}]{miao2023specinfer}
Xupeng Miao, Gabriele Oliaro, Zhihao Zhang, Xinhao Cheng, Zeyu Wang, Zhengxin Zhang, Rae Ying~Yee Wong, Alan Zhu, Lijie Yang, Xiaoxiang Shi, et~al. 2023.
\newblock Specinfer: Accelerating generative large language model serving with tree-based speculative inference and verification.
\newblock \emph{arXiv preprint arXiv:2305.09781}.

\bibitem[{Ni et~al.(2024)Ni, Liu, Tang, Han, and Wang}]{ni2024ems}
Yunsheng Ni, Chuanjian Liu, Yehui Tang, Kai Han, and Yunhe Wang. 2024.
\newblock Ems-sd: Efficient multi-sample speculative decoding for accelerating large language models.
\newblock \emph{arXiv preprint arXiv:2405.07542}.

\bibitem[{Qin et~al.(2024)Qin, Sun, Li, Shen, Sun, and Zhong}]{qin2024lightning}
Zhen Qin, Weigao Sun, Dong Li, Xuyang Shen, Weixuan Sun, and Yiran Zhong. 2024.
\newblock Lightning attention-2: A free lunch for handling unlimited sequence lengths in large language models.
\newblock \emph{arXiv preprint arXiv:2401.04658}.

\bibitem[{Reimers and Gurevych(2019)}]{reimers-gurevych-2019-sentence}
Nils Reimers and Iryna Gurevych. 2019.
\newblock \href {https://doi.org/10.18653/v1/D19-1410} {Sentence-{BERT}: Sentence embeddings using {S}iamese {BERT}-networks}.
\newblock In \emph{Proceedings of the 2019 Conference on Empirical Methods in Natural Language Processing and the 9th International Joint Conference on Natural Language Processing (EMNLP-IJCNLP)}, pages 3982--3992, Hong Kong, China. Association for Computational Linguistics.

\bibitem[{Ryu and Kim(2024)}]{ryu2024closer}
Hyun Ryu and Eric Kim. 2024.
\newblock Closer look at efficient inference methods: A survey of speculative decoding.
\newblock \emph{arXiv preprint arXiv:2411.13157}.

\bibitem[{Saxena(2023)}]{saxena2023prompt}
Apoorv Saxena. 2023.
\newblock \href {https://github.com/apoorvumang/prompt-lookup-decoding/} {Prompt lookup decoding}.

\bibitem[{Shah et~al.(2024)Shah, Bikshandi, Zhang, Thakkar, Ramani, and Dao}]{shah2024flashattention}
Jay Shah, Ganesh Bikshandi, Ying Zhang, Vijay Thakkar, Pradeep Ramani, and Tri Dao. 2024.
\newblock Flashattention-3: Fast and accurate attention with asynchrony and low-precision.
\newblock \emph{arXiv preprint arXiv:2407.08608}.

\bibitem[{Shlens(2014)}]{shlens2014tutorial}
Jonathon Shlens. 2014.
\newblock A tutorial on principal component analysis.
\newblock \emph{arXiv preprint arXiv:1404.1100}.

\bibitem[{Shoeybi et~al.(2019)Shoeybi, Patwary, Puri, LeGresley, Casper, and Catanzaro}]{shoeybi2019megatron}
Mohammad Shoeybi, Mostofa Patwary, Raul Puri, Patrick LeGresley, Jared Casper, and Bryan Catanzaro. 2019.
\newblock Megatron-lm: Training multi-billion parameter language models using model parallelism.
\newblock \emph{arXiv preprint arXiv:1909.08053}.

\bibitem[{Shrivastava and Li(2014)}]{shrivastava2014asymmetric}
Anshumali Shrivastava and Ping Li. 2014.
\newblock Asymmetric lsh (alsh) for sublinear time maximum inner product search (mips).
\newblock \emph{Advances in neural information processing systems}, 27.

\bibitem[{Stern et~al.(2018)Stern, Shazeer, and Uszkoreit}]{stern2018blockwise}
Mitchell Stern, Noam Shazeer, and Jakob Uszkoreit. 2018.
\newblock Blockwise parallel decoding for deep autoregressive models.
\newblock \emph{Advances in Neural Information Processing Systems}, 31.

\bibitem[{Stewart et~al.(2024)Stewart, Trager, Gonugondla, and Soatto}]{stewart2024n}
Lawrence Stewart, Matthew Trager, Sujan~Kumar Gonugondla, and Stefano Soatto. 2024.
\newblock The n-grammys: Accelerating autoregressive inference with learning-free batched speculation.
\newblock \emph{arXiv preprint arXiv:2411.03786}.

\bibitem[{Sun et~al.(2023)Sun, Simcha, Dopson, Guo, and Kumar}]{soar_2023}
Philip Sun, David Simcha, Dave Dopson, Ruiqi Guo, and Sanjiv Kumar. 2023.
\newblock \href {https://arxiv.org/abs/2404.00774} {Soar: Improved indexing for approximate nearest neighbor search}.
\newblock In \emph{Neural Information Processing Systems}.

\bibitem[{Touvron et~al.(2023)Touvron, Martin, Stone, Albert, Almahairi, Babaei, Bashlykov, Batra, Bhargava, Bhosale et~al.}]{touvron2023llama}
Hugo Touvron, Louis Martin, Kevin Stone, Peter Albert, Amjad Almahairi, Yasmine Babaei, Nikolay Bashlykov, Soumya Batra, Prajjwal Bhargava, Shruti Bhosale, et~al. 2023.
\newblock Llama 2: Open foundation and fine-tuned chat models.
\newblock \emph{arXiv preprint arXiv:2307.09288}.

\bibitem[{Vaswani(2017)}]{vaswani2017attention}
A~Vaswani. 2017.
\newblock Attention is all you need.
\newblock \emph{Advances in Neural Information Processing Systems}.

\bibitem[{Wolf(2019)}]{wolf2019huggingface}
T~Wolf. 2019.
\newblock Huggingface's transformers: State-of-the-art natural language processing.
\newblock \emph{arXiv preprint arXiv:1910.03771}.

\bibitem[{Xia et~al.(2024)Xia, Yang, Dong, Wang, Li, Ge, Liu, Li, and Sui}]{xia2024unlocking}
Heming Xia, Zhe Yang, Qingxiu Dong, Peiyi Wang, Yongqi Li, Tao Ge, Tianyu Liu, Wenjie Li, and Zhifang Sui. 2024.
\newblock Unlocking efficiency in large language model inference: A comprehensive survey of speculative decoding.
\newblock \emph{arXiv preprint arXiv:2401.07851}.

\bibitem[{Yang et~al.(2023)Yang, Ge, Wang, Jiao, Jiang, Yang, Majumder, and Wei}]{yang2023inference}
Nan Yang, Tao Ge, Liang Wang, Binxing Jiao, Daxin Jiang, Linjun Yang, Rangan Majumder, and Furu Wei. 2023.
\newblock Inference with reference: Lossless acceleration of large language models.
\newblock \emph{arXiv preprint arXiv:2304.04487}.

\bibitem[{Yang et~al.(2024)Yang, Huang, Dai, and Chen}]{yang2024multi}
Sen Yang, Shujian Huang, Xinyu Dai, and Jiajun Chen. 2024.
\newblock Multi-candidate speculative decoding.
\newblock \emph{arXiv preprint arXiv:2401.06706}.

\bibitem[{Zhang et~al.(2024{\natexlab{a}})Zhang, Liu, and Song}]{zhang2024beyond}
Chen Zhang, Zhuorui Liu, and Dawei Song. 2024{\natexlab{a}}.
\newblock Beyond the speculative game: A survey of speculative execution in large language models.
\newblock \emph{arXiv preprint arXiv:2404.14897}.

\bibitem[{Zhang et~al.(2024{\natexlab{b}})Zhang, Zhu, Yang, Xu, Li, Phothilimthana, and Jia}]{zhang2024accelerating}
Zhihao Zhang, Alan Zhu, Lijie Yang, Yihua Xu, Lanting Li, Phitchaya~Mangpo Phothilimthana, and Zhihao Jia. 2024{\natexlab{b}}.
\newblock Accelerating retrieval-augmented language model serving with speculation.
\newblock \emph{arXiv preprint arXiv:2401.14021}.

\bibitem[{Zheng et~al.(2023)Zheng, Chiang, Sheng, Zhuang, Wu, Zhuang, Lin, Li, Li, Xing et~al.}]{zheng2023judging}
Lianmin Zheng, Wei-Lin Chiang, Ying Sheng, Siyuan Zhuang, Zhanghao Wu, Yonghao Zhuang, Zi~Lin, Zhuohan Li, Dacheng Li, Eric Xing, et~al. 2023.
\newblock Judging llm-as-a-judge with mt-bench and chatbot arena.
\newblock \emph{Advances in Neural Information Processing Systems}, 36:46595--46623.

\bibitem[{Zimmer et~al.(2024)Zimmer, Gritta, Lampouras, Ammar, and Wang}]{zimmer2024mixture}
Matthieu Zimmer, Milan Gritta, Gerasimos Lampouras, Haitham~Bou Ammar, and Jun Wang. 2024.
\newblock Mixture of attentions for speculative decoding.
\newblock \emph{arXiv preprint arXiv:2410.03804}.

\end{thebibliography}

\appendix

\section{Appendix}
\label{sec:appendix}

The following tables contain the results used to generate the bar charts in the paper.

\begin{table}[thbp]
\centering
 \resizebox{\linewidth}{!}{\begin{tabular}{lcccc}
 \toprule
TPS & CodeLlama-7B & Llama2-Chat-7B & CodeLlama-13B & Llama2-Chat-13B\\ \midrule
LLM & 19 & 19 & 15 & 15\\
SD & 31 & 25 & 21 & 20\\
PLD & 42 & 21 & 29 & 19\\
REST & 72 & 55 & 39 & 32\\
DReSD & 104 & 78 & 85 & 50\\
 \bottomrule
 \end{tabular}}
 \caption{Fastest configurations for a selection of methods (greedy decoding), relative to auto-regressive generation (LLM), CL = CodeLlama, LC = Llama2-Chat.}
 \label{table:fig1}
\end{table}

\begin{table}[thbp]
\centering
 \resizebox{\linewidth}{!}{\begin{tabular}{lcccc}
 \toprule
MAR & CodeLlama-7B & Llama2-Chat-7B & CodeLlama-13B & Llama2-Chat-13B \\ \midrule
SD & 29.6 & 17.8 & 18.9 & 19.1\\
PLD & 32.7 & 10.0 & 22.4 & 12.7\\
REST & 28.1 & 18.0 & 20.3 & 17.9\\
DReSD & 45.5 & 31.9 & 32.4 & 32.9\\
REST-I & 33.4 & 35.0 & 27.0 & 29.5\\
DReSD-I & 49.4 & 48.1 & 51.3 & 46.3\\ \bottomrule
 \end{tabular}}
 \caption{Mean Acceptance Rates (MAR) for the Code
Assistant. Suffix "-I" denotes the ID datastore setting.}
 \label{table:fig5}
\end{table}

\begin{table}[thbp]
\centering
 \resizebox{\linewidth}{!}{\begin{tabular}{lcccccccccc}
 \toprule
TPS & CodeLlama-7B & Llama2-Chat-7B & CodeLlama-13B & Llama2-Chat-13B\\ \midrule
REST-G & 32 & 27 & 19 & 17\\
DReSD-G & 37 & 32 & 21 & 22\\
REST-IG & 72 & 55 & 39 & 32\\
DReSD-IG & 75 & 61 & 47 & 36\\
DReSD-IBG & 104 & 78 & 85 & 50\\ \midrule
REST-N & 34 & 28 & 19 & 18\\
DReSD-N & 39 & 36 & 22 & 22\\
REST-IN & 45 & 48 & 26 & 27\\
DReSD-IN & 48 & 49 & 30 & 30\\
DReSD-IBN & 52 & 53 & 32 & 33\\ \bottomrule
 \end{tabular}}
 \caption{Tokens-per-second for a selection of LLMs and configurations: "-G" = greedy decoding, "-I" = uses the
ID datastore, "-N" = nucleus sampling, "-B" = our best setup (see section 5.5), LLM = auto-regressive generation.}
 \label{table:fig6}
\end{table}

\begin{table}[thbp]
\centering
  \resizebox{\linewidth}{!}{\begin{tabular}{lcc}
 \toprule
MAR & CodeLlama-Instruct-7B & Llama2-Chat-7B \\ \midrule
REST-C & 16.7 & 18.0\\
REST-M & 9.3 & 8.0\\ \midrule
DReSD-C & 29.0 & 31.9\\
DReSD-M & 18.5 & 17.3\\ \bottomrule
 \end{tabular}}
 \caption{Mean Acceptance Rates with high (CodeAl-
paca, "-C") \& low (MT-Bench, "-M") prompt alignment.}
 \label{table:fig7}
\end{table}

\begin{table}[thbp]
\centering
  \resizebox{\linewidth}{!}{\begin{tabular}{lcccc}
 \toprule
MAR & CodeLlama-Instruct-7B & Llama2-Chat-7B \\ \midrule
REST-IG & 10.0 & 8.4\\
DReSD-IG & 25.2 & 22.4\\ \bottomrule
REST-IN & 11.0 & 9.4\\
DReSD-IN & 25.3 & 23.4\\ \bottomrule
 \end{tabular}}
 \caption{Mean Accepted Rates with an ID datastore
"-I", nucleus sampling "-N" and greedy "-G" decoding.}
 \label{table:fig8}
\end{table}

\end{document}